\documentclass[11pt]{article}
\usepackage{amsfonts, amsmath, amssymb, amsthm, float, fancyhdr}
\usepackage[pdftex]{graphicx}
\usepackage[margin=1in]{geometry}
\usepackage{enumerate}
\usepackage{tikz}
\usepackage{textcomp}
\usetikzlibrary{positioning,automata}

\makeatletter
\renewcommand*\env@matrix[1][*\c@MaxMatrixCols c]{%
\hskip -\arraycolsep
\let\@ifnextchar\new@ifnextchar
\array{#1}}
\makeatother

\everymath{\displaystyle}
\DeclareGraphicsExtensions{.pdf,.png,.jpg}

\begin{document}
\begin{center}
\huge{On the Computational Power of RNNs}\\

\bigskip

\bigskip

\large{Samuel A. Korsky, Robert C. Berwick}\\

\bigskip

\bigskip

May 2019
\end{center}
\bigskip

\bigskip
\noindent
\begin{center}
{\bf{{\large{ABSTRACT}}}}\\
\end{center}
\noindent
Recent neural network architectures such as the basic recurrent neural network (RNN) and Gated Recurrent Unit (GRU) have gained prominence as end-to-end learning architectures for natural language processing tasks.  But what is the computational power of such systems? We prove that finite precision RNNs with one hidden layer and ReLU activation and finite precision GRUs are exactly as computationally powerful as deterministic finite automata. Allowing arbitrary precision, we prove that RNNs with one hidden layer and ReLU activation are at least as computationally powerful as pushdown automata. If we also allow infinite precision, infinite edge weights, and nonlinear output activation functions, we prove that GRUs are at least as computationally powerful as pushdown automata. All results are shown constructively.

\section*{Introduction}
Recent work [1] suggests that recurrent ``neural network" models of several types perform better than sequential models in acquiring and processing hierarchical structure. Indeed, recurrent networks have achieved state-of-the-art results in a number of natural language processing tasks, including named-entity recognition [2], language modeling [3], sentiment analysis [4], natural language generation [5], and beyond.

\bigskip
\noindent
The hierarchical structure associated with natural languages is often modeled as some variant of context-free languages, whose languages may be defined over an alphabet $\Sigma$.  These context-free languages are exactly those that can be recognized by pushdown automata (PDAs). Thus it is natural to ask whether these modern natural language processing tools, including simple recurrent neural networks (RNNs) and other, more advanced recurrent architectures, can learn to recognize these languages.

\bigskip
\noindent
The computational power of RNNs has been studied extensively using empirical testing. Much of this research [8], [9] focused on the ability of RNNs to recognize simple context-free languages such as $a^nb^n$ and $a^nb^mB^mA^n$, or context-sensitive languages such as $a^nb^nc^n$. Related works [10], [11], [12] focus instead on Dyck languages of balanced parenthesis, which motivates some of our methods. Gated architectures such as the Gated Recurrent Unit (GRU) and Long Short-Term Memory (LSTM) obtain high accuracies on each of these tasks. While simpler RNNs have also been tested, one difficulty is that the standard hyperbolic tangent activation function makes counting difficult. On the other hand, RNNs with ReLU activations were found to perform better, but suffer from what is known as the ``exploding gradient problem" and thus are more difficult to train [8]. 

\bigskip
\noindent
Instead of focusing on a single task, many researchers have studied the broader {\it{theoretical}} computational power of recurrent models, where weights are not trained but rather initialized to recognize a desired language. A celebrated result [6] shows that a simple recurrent architecture with 1058 hidden nodes and a saturated-linear activation $\sigma$ is a universal Turing Machine, with:
$$ \sigma(x) = \begin{cases}0, & x < 0\\x, & 0 \le x \le 1\\1, & x > 1\end{cases} $$
However, their architecture encodes the whole input in its internal state and the relevant computation is only performed after reading a terminal token. This differs from more common RNN variants that consume tokenized inputs at each time step. Furthermore, the authors admit that were the saturated-linear activation to be replaced with the similar and more common sigmoid or hyperbolic tangent activation functions, their methodology would fail. 

\bigskip
\noindent
More recent work [7] suggests that single-layer RNNs with rectified linear unit (ReLU) activations and softmax outputs can also be simulated as universal Turing Machines, but this approach again suffers from the assumption that the entire input is read before computation occurs. 

\bigskip
\noindent
Motivated by these earlier theoretical results, in this report we seek to show results about the computational power of recurrent architectures actually used in practice - namely, those that read tokens one at a time and that use standard rather than specially chosen activation functions. In particular we will prove that, allowing infinite precision, RNNs with just one hidden layer and ReLU activation are at least as powerful as PDAs, and that GRUs are at least as powerful as deterministic finite automata (DFAs). Furthermore, we show that using infinite edge weights and a non-standard output function, GRUs are also at least as powerful as PDAs. 

\section{Simple RNNs}
Let a {\it{simple RNN}} be an RNN with the following architecture:
\begin{align*}
h_t &= f(W_xx_t + W_hh_{t - 1} + b_h)\\
o_t &= W_oh_t + b_o
\end{align*}
where $o_i \in \mathbb{R}$ for all $i$, for some chosen activation function $f$, usually the ReLU or the hyperbolic tangent functions. We assume that the inputs are {\it{one-hots}} of a given set of symbols $\Sigma$, vectors of length $|\Sigma|$ where each element but one is $0$ and the remaining element is $1$. 

\bigskip
\noindent
Say that an RNN {\it{accepts}} an input $w$ of length $n$ if after passing $w$ through the RNN, its final output $o_n$ belongs to a predetermined set $S \subseteq \mathbb{R}$, for which membership can be tested in $O(1)$ time. Let the $S$-{\it{language}} of an RNN consist exactly of all inputs that it accepts given set $S$. 

\bigskip
\noindent
In practice, the inputs and hidden nodes of an RNN are stored as numbers with finite precision. Including this restriction, we show the following result:

\bigskip
\noindent
{\bf{Theorem 1.1}}. For every language $L \subseteq \Sigma^*$, $L$ is regular if and only if $L$ is the $S$-language of some finite precision simple RNN. 

\bigskip
\noindent
{\it{Proof.}} We begin with the ``if" direction. Suppose we are given some simple RNN and set $S \subseteq \mathbb{R}$. It suffices to show that there exists a DFA that accepts the $S$-language of this RNN. Assume that the RNN has $m$ hidden nodes, and that these hidden nodes are precise up to $k$ bits. Then there are exactly $2^{mk}$ possible hidden states for the RNN. Construct the following DFA with: 
\begin{itemize}
\item
set of $2^{mk}$ states $Q = \{q_h:h\ \text{is a possible hidden state of the RNN}\}$
\item
alphabet $\Sigma$
\item
transition function $\delta$ where $\delta(q_h, x) = q_{f(W_xx + W_hh + b_h)}$
\item
initial state $q_{h_0}$
\item
set of accepting states $F = \{q_h|W_hh + b_o \in S\}$
\end{itemize} 
It's clear that after reading the first $n$ inputs of a word $w$, the current state of this DFA is $q_{h_n}$, which immediately completes the proof of this direction.

\bigskip
\noindent
For the ``only if" direction, suppose we have a DFA $D = (Q, \Sigma, \delta, q_0, F)$ with corresponding language $L$. We will construct a simple RNN whose inputs are one-hotted symbols from $\Sigma$, with ReLU activation function $f(x) = \text{max}(0, x)$, and with $|Q||\Sigma|$ hidden nodes whose $\{0\}$-language is $L$.

\bigskip
\noindent
The RNN has three layers: the first layer (input layer) has $|\Sigma| + |Q||\Sigma|$ nodes; the second layer (hidden layer) has $|Q||\Sigma|$ nodes; and the third layer (output layer) has one node. For the $|\Sigma|$ nodes in the input layer associated with the one-hot of the current symbol, label each node with its corresponding symbol from $\Sigma$. Label the $|Q||\Sigma|$ hidden nodes (in both the first and second layers) with all $|Q||\Sigma|$ symbol-state combinations $(x, q)$ for $x \in \Sigma$ and $q \in Q$. 

\bigskip
\noindent
For every $x \in \Sigma$, connect the node in the input layer with label $x$ to all nodes in the hidden layer with labels $(x, q)$ for any $q \in Q$ with edges with weight $1$. For all $(x, q) \in \Sigma \times Q$, connect the node in the input layer with label $(x, q)$ to all nodes in the hidden layer with labels $(x', q')$ where $\delta(q, x') = q'$ with edges also of weight $1$. Finally, for all $(x, q) \in \Sigma \times Q/F$, connect the node in the hidden layer with label $(x, q)$ to the single node in the output layer with an edge of weight $1$. 

\bigskip
\noindent
Each of the hidden nodes are initialized to $0$ except a single hidden node with label $(x, q_0)$ for a randomly chosen $x \in \Sigma$, which is initialized to $1$. To complete the description of the RNN, we set $b_h = -1$ and $b_o = 0$. We claim that the following invariant is maintained: after reading some word, suppose the current state of $D$ is $q$. Then after reading the same word, the hidden nodes of the RNN would all be equal to $0$ except for one node with label $(x, q)$ for some $x \in \Sigma$, which would equal $1$. 

\bigskip
\noindent
We prove the claim by induction on the length of the inputted word $n$. The base case of $n = 0$ is trivial. Now assume that after reading a word of length $n$ the current state of $D$ is $q$, and after reading that same word all hidden nodes of the RNN are equal to $0$ except one node with label $(x, q)$ for some $x \in \Sigma$, which is equal to $1$. If the next symbol is $x'$, then the current state of $D$ would be $q'$ where $\delta(q, x') = q'$. For the RNN, the input layer will have exactly two $1$s, namely the node with label $x'$ and the node with label $(x, q)$. Since all edges have weight $1$, that means that before adding $b_h$ or applying $f$ the maximum value a node in the hidden layer can take on is $2$. For this to occur it must be connected to both the nodes in the input layer with value $1$, and thus by definition its label must be $(x', \delta(q, x')) = (x', q')$. By integrality every other node in the hidden layer will take on a value of at most $1$, so after adding $b_h = -1$ and applying $f$ we easily see that the invariant is maintained.

\bigskip
\noindent
Utilizing this invariant it is clear that upon reading a word $w \in L$ the RNN will output $0$, and upon reading a word $w \not\in L$ it will output $1$. Thus $L$ is precisely the $\{0\}$-language of the RNN and the theorem is proven.\qed

\bigskip
\noindent
{\bf{Discussion 1.2}}. This result shows that simple RNNs with finite precision are exactly as computationally powerful as DFAs. In terms of reducing the size of the hidden layer constructed in the proof of the ``only if" direction, it seems likely that $|Q||\Sigma|$ is optimal since $\delta$ is defined on $|Q||\Sigma|$ inputs and needs to be captured fully by the RNN.

\bigskip
\noindent
Removing the finite precision stipulation unsurprisingly increases the capabilities of RNNs. It is natural to now ask whether these simple RNNs can recognize more complicated $S$-languages, and indeed the answer is affirmative. Thus we shift our focus to context-free languages. We begin with some preliminaries:

\bigskip
\noindent
The {\it{Dyck language}} $D_n$ consists of all words over the size $2n$ alphabet $\Sigma = \bigcup\limits_{i = 1}^{n}\{(_i, )_i\}$ that correspond to a balanced string of $n$ types of parentheses. We also define the set of proper prefixes
$$ P_n = \{w|w \not\in D_n, \exists v \in \Sigma^*\ \text{such that}\ wv \in D_n\} $$
so that any word in $P_n$ is the prefix of a word in $D_n$ but is itself unbalanced. We proceed with a motivating theorem:

\bigskip
\noindent
{\bf{Theorem 1.3}} (Chomsky-Sch$\ddot{\text{u}}$tzenberger Theorem). Any context-free language $L$ can be written as $L = D_n \cap R$ for some $n \in \mathbb{Z}^{+}$ and regular language $R$ after a suitable relabeling.

\bigskip
\noindent
{\it{Proof.}} The interested reader may find a proof in [13]. \qed

\bigskip
\noindent
Thus it makes sense to focus on constructing sets $S$ and simple RNNs whose $S$-language is $D_n$. Indeed, since $D_n = g^{-1}(D_2)$ for some homomorphism $g$, we start by focusing on $D_2$, in some sense the ``hardest" context-free language.

\bigskip
\noindent
The critical idea is to ``memorize" an input in the binary representation of some rational number, simulating a stack. Indeed, consider associating with any word $w = w_1w_2\dots w_m \in D_2$ a state $s \in \mathbb{Q}$, defined as follows:
\begin{align*}
s_0 &= 0\\
s_t &= \begin{cases}s_{t - 1}/2, & w_t = (_1\\s_{t - 1}/2 + 1/2, & w_t = (_2\\2s_{t - 1}, & w_t =\ )_1\\2s_{t - 1} - 1, & w_t =\ )_2\end{cases}
\end{align*}
Consider the word $(_2(_1)_1(_2(_1)_1)_2)_2$. The evolution of the state as the word is read symbol by symbol is given by
$$ 0, 0.1, 0.01, 0.1, 0.11, 0.011, 0.11, 0.1, 0 $$
This example makes it clear that this notion of state accurately captures all the relevant information about words in $P_2 \cup D_2$. 

\bigskip
\noindent
The difficulty in capturing this notion of state in a RNN is that the constant to multiply $s_{t - 1}$ by changes depending on the input (it can be either $2$ or $1/2$ in our example above). Thus storing $s_t$ in a single hidden node is impossible. Instead, we use two hidden nodes. Below, we generalize from $D_2$ to $D_n$. 

\bigskip
\noindent
Ignoring the output layer for now, consider the simple RNN defined by
\begin{align*}
h_0 &= [0\ \ 0]^T\\
h_t &= \text{ReLU}(W_xx_t + W_hh_{t - 1})
\end{align*}
where the inputs $x$ are $2n \times 1$ one-hots of the symbols in $\Sigma$ (the alphabet of $D_n$) in the order $(_1, (_2, \dots, (_n, )_1, )_2, \dots, )_n$ and the hidden states have dimension $2 \times 1$ where
\begin{align*}
W_{x} &= \begin{bmatrix}2(2n + 1)^{-1} & 4(2n + 1)^{-1} & \dots & 2n(2n + 1)^{-1} & -2n - 1 & -2n - 1 & \dots & -2n - 1\\-2n - 1 & -2n - 1 & \dots & -2n - 1 & -2 & -4 & \dots & -2n\end{bmatrix}\\
\\
W_{h} &= \begin{bmatrix}(2n + 1)^{-1} & (2n + 1)^{-1}\\2n + 1 & 2n + 1\end{bmatrix}
\end{align*}
As before, associate with each word $w = w_1w_2\dots w_m \in D_n$ a state $s \in \mathbb{Q}$ now satisfying
\begin{align*}
s_0 &= 0\\
s_t &= \begin{cases}(2n + 1)^{-1}s_{t - 1} + 2i(2n + 1)^{-1}, & w_t =(_i \\(2n + 1)s_{t - 1} - 2i, & w_t =\ )_i\end{cases}
\end{align*}
for all $i \in \{1, 2, \dots, n\}$. 

\bigskip
\noindent
This is similar to the state we defined before, though now generalized to $D_n$ and also with intentionally present blank space inserted between the digits in base $2n + 1$. We will show the following invariant:

\bigskip
\noindent
{\bf{Lemma 1.4}}. Given an input word $w \in P_n \cup D_n$, we have $h_t = [s_t\ \ 0]^T$ or $h_t = [0\ \ s_t]^T$ for all $t$. 

\bigskip
\noindent
{\it{Proof}}. We proceed by induction on $t$. The base case of $t = 0$ is trivial. Now, suppose $w_{t + 1} = (_i$ for some $i \in \{1, 2, \dots, n\}$ and assume without loss of generality that $h_t = [s_t\ \ 0]^T$. Then
$$ h_{t + 1} = \text{ReLU}(W_xx_{t + 1} + W_hh_t) = \text{ReLU}\left([2i(2n + 1)^{-1}\ \ -2n - 1]^T + [(2n + 1)^{-1}s_t\ \ (2n + 1)s_t]^T\right) $$
Now, since $w \in P_n \cup D_n$ we have that $s_t \in [0, 1)$ for any $t$, which follows immediately from the stack interpretation of the base $2n + 1$ representation of $s_t$. Thus $\text{ReLU}(-2n - 1 + (2n + 1)s_t) = 0$ and so 
$$ h_{t + 1} = [(2n + 1)^{-1}s_t + 2i(2n + 1)^{-1}\ \ 0]^T = [s_{t + 1}\ \ 0]^T $$
as desired. Alternatively, suppose $w_{t + 1} = )_i$ for some $i \in \{1, 2, \dots, n\}$. Again, assume without loss of generality that $h_t = [s_t\ \ 0]^T$. Then
$$ h_{t + 1} = \text{ReLU}(W_xx_{t + 1} + W_hh_t) = \text{ReLU}\left([-2n - 1\ \ -2i]^T + [(2n + 1)^{-1}s_t\ \ (2n + 1)s_t]^T\right) $$
The fact that $w \in P_n \cup D_n$ clearly implies that $(2n + 1)s_t - 2i \ge 0$ and so we have that
$$ h_{t + 1} = [0\ \ (2n + 1)s_t - 2i]^T = [0\ \ s_{t + 1}]^T $$
which completes the induction.\qed

\bigskip
\noindent
A pictorial example of this RNN is depicted below for $n = 2$: 

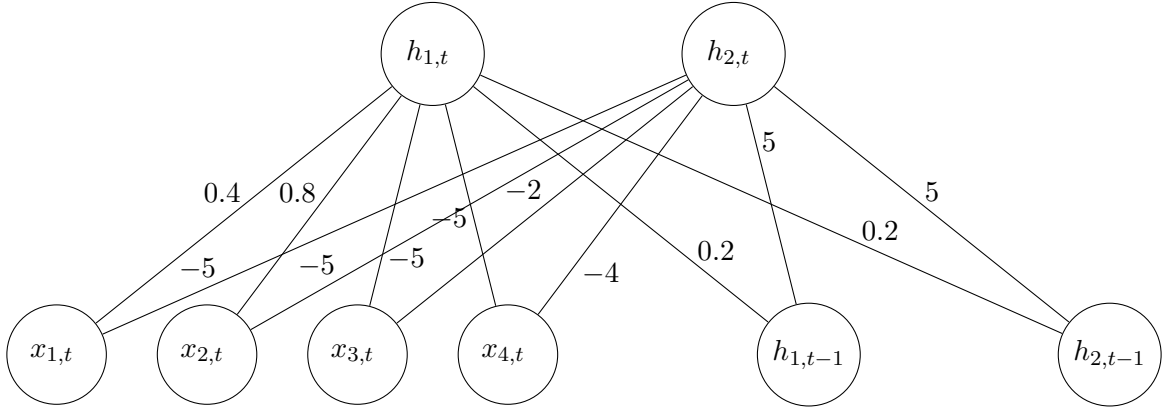
\begin{center}
\tikzstyle{vertex}=[circle, draw]
\begin{tikzpicture}[transform shape]
\node[vertex](r1) at (-2, 2) {\ $ h_{1, t}\ \ $};
\node[vertex](r2) at (2, 2) {$ \ h_{2, t}\ \ $};
\node[vertex](q1) at (-7,-2) {$\ x_{1, t}\ \ $};
\node[vertex](q2) at (-5,-2) {$\ x_{2, t}\ \ $};
\node[vertex](q3) at (-3,-2) {$\ x_{3, t}\ \ $};
\node[vertex](q4) at (-1,-2) {$\ x_{4, t}\ \ $};
\node[vertex](h1) at (3,-2) {$ h_{1, t - 1} $};
\node[vertex](h2) at (7,-2) {$ h_{2, t - 1} $};
\begin{scope}[every path/.style={-}, every node/.style={inner sep=1pt}]
\draw (r1) -- node [pos=0.5, anchor=south east] {$0.4$} (q1);
\draw (r1) -- node [pos=0.5, anchor=south east] {$0.8$} (q2);
\draw (r1) -- node [pos=0.7, anchor=north west] {$-5$} (q3);
\draw (r1) -- node [pos=0.5, anchor=north east] {$-5$} (q4);
\draw (r1) -- node [pos=0.75, anchor=south west] {$0.2$} (h1);
\draw (r1) -- node [pos=0.65, anchor=south west] {$0.2$} (h2);
\draw (r2) -- node [anchor=south east, pos=0.8] {$-5$} (q1);
\draw (r2) -- node [anchor=south east, pos=0.8] {$-5$} (q2);
\draw (r2) -- node [pos=0.5, anchor=south east] {$-2$} (q3);
\draw (r2) -- node [pos=0.75, anchor=north west] {$-4$} (q4);
\draw (r2) -- node [pos=0.25, anchor=south west] {$5$} (h1);
\draw (r2) -- node [pos=0.5, anchor=south west] {$5$} (h2);
\end{scope} 
\end{tikzpicture}
\end{center}

\bigskip
\noindent
Thus we have found an efficient way to store $s_t$. Now it's clear that for any $w = w_1w_2\dots w_m \in P_n$ we have $s_m > 0$ and for any $w = w_1w_2\dots w_m \in D_n$ we have $s_m = 0$, so it is tempting to try and add a simple output layer to this RNN and claim that its $\{0\}$-language is $D_n$. However, this is most likely impossible to accomplish. 

\bigskip
\noindent
Indeed, consider the word $w = )_1(_1$. We have that $s_2 = 0$ for this word, but $w \not\in D_n$. Furthermore, consider the word $w = (_2)_1(_1)_2$. We have that $s_t \ge 0$ for all $t$ and $s_4 = 0$ for this word, yet $w \not\in D_n$. Hence we must be able to flag when an inappropriate closing parenthesis appears in an input and retain that information while reading the rest of the input. To that end, consider the following simple RNN, an example of which can be found in Appendix A.1:
\begin{align*}
h_0 &= [0\ \ 0\ \ 0\ \ 0\ \ 0\ \ 0]^T\\
h_t &= \text{ReLU}(W_xx_t + W_hh_{t - 1})\\
o_t &= W_oh_t
\end{align*}
where again the inputs $x$ are $2n \times 1$ one-hots of the symbols in $\Sigma$ (the alphabet of $D_n$) in the order $(_1, (_2, \dots, (_n, )_1, )_2, \dots, )_n$ and the hidden states have dimension $6 \times 1$ where
\begin{align*}
W_{x} &= \begin{bmatrix}2(2n + 1)^{-1} & 4(2n + 1)^{-1} & \dots & 2n(2n + 1)^{-1} & -2n - 1 & -2n - 1 & \dots & -2n - 1\\-2n - 1 & -2n - 1 & \dots & -2n - 1 & -2 & -4 & \dots & -2n\\0 & 0 & \dots & 0 & 2 & 4 & \dots & 2n\\0 & 0 & \dots & \dots & \dots & \dots & \dots & 0\\-2n - 1 & -2n - 1 & \dots & -2n - 1 & -3 & -5 & \dots & -2n - 1\\0 & 0 & \dots & \dots & \dots & \dots & \dots & 0\end{bmatrix}\\
\\
W_{h} &= \begin{bmatrix}(2n + 1)^{-1} & (2n + 1)^{-1} & 0 & 0 & 0 & 0\\2n + 1 & 2n + 1 & 0 & 0 & 0 & 0\\-2n - 1 & -2n - 1 & 0 & 0 & 0 & 0\\0 & 0 & 1 & 1 & 0 & 0\\2n + 1 & 2n + 1 & 0 & 0 & 0 & 0\\0 & 0 & 0 & 0 & 1 & 1\end{bmatrix}\\
\\
W_o &= [1\ \ 1\ \ 1\ \ 1\ \ 1\ \ 1]^T
\end{align*}
Because the last four elements of the first two rows of $W_h$ are all equal to $0$ and otherwise the first two rows of $W_x$ and $W_h$ are the same as before, it is clear that Lemma 1.4 still applies in some form for the new simple RNN. Indeed, denoting
$$ h_t = [h_{1, t}\ \ h_{2, t}\ \ h_{3, t}\ \ h_{4, t}\ \ h_{5, t}\ \ h_{6, t}]^T $$ 
we have

\bigskip
\noindent
{\bf{Corollary 1.5}}. With respect to a word $w \in P_n \cup D_n$, we have $[h_{1, t}\ \ h_{2, t}] = [s_t\ \ 0]$ or $[h_{1, t}\ \ h_{2, t}] = [0\ \ s_t]$ for all $t$.

\bigskip
\noindent
We proceed with an important lemma:

\bigskip
\noindent
{\bf{Lemma 1.6}}. For any word $w \in P_n$, there is a unique $x \in \{)_1, )_2, \dots, )_n\}$ such that $wx \in P_n \cup D_n$.

\bigskip
\noindent
{\it{Proof}}. This immediately follows from the definition of a balanced string. Indeed, if $s$ is the state associated with $w$ then this unique $x$ is given by
$$ x = )_i \Longleftrightarrow \frac{2i}{2n + 1} \le s < \frac{2i + 1}{2n + 1} $$\qed

\bigskip
\noindent
We are now ready to show the following:

\bigskip
\noindent
{\bf{Lemma 1.7}}. Given an input word $w = w_1w_2\dots w_m \in P_n \cup D_n$, we have that $h_{3, m} = h_{5, m} = 0$.

\bigskip
\noindent
{\it{Proof}}. We first restrict our attention to $h_{3, m}$. Note that
$$ h_{3, m} = \begin{cases}\text{ReLU}\left(-(2n + 1)h_{1, m-1} - (2n + 1)h_{2, m-1}\right), & w_m = (_i\\\text{ReLU}\left(2i - (2n + 1)h_{1, m-1} - (2n + 1)h_{2, m-1}\right), & w_m =\ )_i\end{cases} $$
for any $i$, which follows from the definition of $W_h$ and $W_x$. Then using Corollary 1.5 we find 
$$ h_{3, m} = \begin{cases}\text{ReLU}\left(-(2n + 1)s_{m - 1}\right), & w_m = (_i\\\text{ReLU}\left(2i - (2n + 1)s_{m-1}\right), & w_m =\ )_i\end{cases} $$
Now using the inequality in the proof of Lemma 1.6 we immediately obtain $h_{3, m} =0$ as desired.

\bigskip
\noindent
Considering now $h_{5, m}$ we notice
$$ h_{5, m} = \begin{cases}\text{ReLU}\left(-2n - 1 + (2n + 1)h_{1, m-1} + (2n + 1)h_{2, m-1}\right), & w_m = (_i\\\text{ReLU}\left(-2i - 1 + (2n + 1)h_{1, m-1} + (2n + 1)h_{2, m-1}\right), & w_m =\ )_i\end{cases} $$
and doing an analysis similar to that for $h_{3, m}$, we obtain $h_{5, m} = 0$ as desired.\qed

\bigskip
\noindent
Applying Lemma 1.6 allows us to make the following statement:

\bigskip
\noindent
{\bf{Lemma 1.8}}. Given a word $w = w_1w_2\dots w_m \in P_n$, consider the unique $j \in \{1, 2, \dots, n\}$ such that $w)_j \in P_n \cap D_n$. Then with respect to a word $w)_i$ with $i > j$, we have $h_{3, m + 1} > 0$. Similarly, with respect to a word $w)_i$ with $i < j$, we have $h_{5, m + 1} > 0$. 

\bigskip
\noindent
{\it{Proof}}. First suppose $i > j$. As in the proof of Lemma 1.7, we use
$$ h_{3, m + 1} = \text{ReLU}\left(2i - (2n + 1)h_{1, m} - (2n + 1)h_{2, m} + h_{3, m}\right) = \text{ReLU}\left(2i - (2n + 1)s_m\right) $$
where we again use Corollary 1.5 and the fact that $h_{3, m} = 0$ from Lemma 1.7. But from the proof of Lemma 1.6, since $w)_j \in P_n \cup D_n$ we know that
$$ \frac{2j}{2n + 1} \le s_m < \frac{2j + 1}{2n + 1} $$
and since $i > j$ we have that $2i > 2j + 1$ since $i$ and $j$ are integral. Thus $h_{3, m + 1} > 0$ as desired.

\bigskip
\noindent
Now assume $i < j$. As in the previous case we obtain
$$ h_{5, m + 1} = \text{ReLU}\left(-2i -1 + (2n + 1)h_{1, m} + (2n + 1)h_{2, m} + h_{5, m}\right) = \text{ReLU}\left(-2i - 1 + (2n + 1)s_m\right) $$
again using Corollary 1.5 and Lemma 1.7. And again using the inequality from the proof of Lemma 1.6 and the fact that $i < j$ we obtain $h_{5, m + 1} > 0$, completing the proof.\qed

\bigskip
\noindent
Thus we have constructed the desired ``flags." Indeed, hidden nodes $h_3$ and $h_5$ remain equal to $0$ while the currently read input lies in $P_n \cup D_n$, but one of these nodes becomes positive the moment the currently read input does not lie in this set. 

\bigskip
\noindent
However, there are still difficulties. It is possible for $h_3$ or $h_5$ to become positive and later return to $0$. Indeed, running the simple RNN on the word $w = (_2)_1(_1)_2(_2)_2$, we compute $h_{1, 6} = h_{2, 6} = h_{3, 6} = h_{5,6} = 0$. However, clearly $w \not\in P_n \cup D_n$. Therefore we need to add architecture that retains the information as to whether the hidden nodes $h_3$ or $h_5$ ever become positive, and below we show that hidden nodes $h_4$ and $h_6$ respectively are sufficient.

\bigskip
\noindent
{\bf{Lemma 1.9}}. For any input $w \in \Sigma^*$ we have
$$ h_{4, t} = \sum_{i = 0}^{t - 1}h_{3, i} $$
$$ h_{6, t} = \sum_{i = 0}^{t - 1}h_{5, i} $$

\bigskip
\noindent
{\it{Proof}}. From the definition of $W_x$ and $W_h$ we have
$$ h_{4, t} = \text{ReLU}(h_{3, t - 1} + h_{4, t - 1}) $$
$$ h_{6, t} = \text{ReLU}(h_{5, t - 1} + h_{6, t - 1}) $$
and since $h_{3, t}, h_{5, t} \ge 0$ for all $t$ (because of the ReLU) we immediately have the result by induction or direct expansion.\qed

\bigskip
\noindent
We are now ready to combine these lemmas and accomplish our original goal:

\bigskip
\noindent
{\bf{Theorem 1.10}}. The $\{0\}$-language of the simple RNN described earlier in the section is $D_n$. 

\bigskip
\noindent
{\it{Proof.}} Consider any input $w = w_1w_2\dots w_m \in \Sigma^*$ into the RNN. For the remainder of the proof, remember that $h_{i, t} \ge 0$ for all $i, t$ because of the ReLU activation. We consider three cases:
\begin{itemize}
\item
Case 1: $w \in P_n \cup D_n$. 
\end{itemize}
In this case by Corollary 1.5 we have $h_{1, m} + h_{2, m} = s_m$. Furthermore, by Lemma 1.7 we have $h_{3, m} = h_{5, m} = 0$. By combining Lemmas 1.7 and 1.9, we have $h_{4, m} = h_{6, m} = 0$. Thus $o_m = s_m$ which, given that $w \in P_n \cup D_n$, equals $0$ precisely when $w \in D_n$, by the inequality from the proof of Lemma 1.6.
\begin{itemize}
\item
Case 2: $w \not\in P_n \cup D_n$ and $w_1w_2\dots w_{m - 1} \in P_n \cup D_n$. 
\end{itemize}
In this case we clearly must have $w_m = )_i$ for some $i \in \{1, 2, \dots, n\}$ and thus by Lemma 1.8 we have that either $h_{3, m} > 0$ or $h_{5, m} > 0$, so $o_m > 0$. 
\begin{itemize}
\item
Case 3: $w_1w_2\dots w_k \not\in P_n \cup D_n$ for some $k \in \{1, 2, \dots, m - 1\}$.
\end{itemize}
Suppose $j$ is the minimal index such that $w_1w_2\dots w_j \not\in P_n \cup D_n$. Then by minimality $w_1w_2\dots w_{j - 1} \in P_n \cup D_n$ so again by Lemma 1.8 we have that either $h_{3, j} > 0$ or $h_{5, j} > 0$. But since $j \le k \le m - 1$ by Lemma 1.9 this means that either $h_{4, m} > 0$ or $h_{6, m} > 0$, so $o_m > 0$.

\bigskip
\noindent
Thus $o_m = 0$ if and only if $w \in D_n$, completing the proof of the theorem.\qed

\bigskip
\noindent
Now recall in the proof of Theorem 1.1 we showed that any regular language $R$ was the $\{0\}$-language of some simple RNN, and moreover that for any input not in $R$ the output of that RNN is positive. This allows us to provide a simple proof of the main theorem of this section:

\bigskip
\noindent
{\bf{Theorem 1.11}}. For any context-free language $L$, suppose we relabel and write $L = D_n \cap R$ for some regular language $R$, whose corresponding minimum-size DFA has $r$ states. Then there exists a simple RNN with a hidden layer of size $6 + 2nr$ whose $\{0\}$-language is $L$. 

\bigskip
\noindent
{\it{Proof}}. 
Consider the simple RNN with $R$ as its $\{0\}$-language described in the proof of Theorem 1.1 and the simple RNN with $D_n$ as its $\{0\}$-language constructed to prove Theorem 1.10. Merge the $|\Sigma| = 2n$ nodes in the input layer corresponding to the input and merge the single output nodes of both RNNs. Stack the two hidden layers, and add no new edges. There were $|\Sigma|r = 2nr$ hidden nodes in the first RNN and $6$ in the second, so altogether the new RNN has $6 + 2nr$ hidden nodes.

\bigskip
\noindent
The output of the new RNN is equal to the summed output of the two original RNNs, and from the proofs of Theorems 1.1 and 1.10 these outputs are always nonnegative. Thus the output of the new RNN is $0$ if and only if the outputs of both old RNNs were $0$, immediately proving the theorem.\qed

\bigskip
\noindent
{\bf{Discussion 1.12}}. This result shows that simple RNNs with arbitrary precision are at least as computationally powerful as PDAs. 

\section{Gated RNNs}
In practice, architectures more complicated than the simple RNNs studied above - notably gated RNNs, including the Gated Recurrent Unit (GRU) and Long Short-Term Memory (LSTM) - perform better on many natural language tasks. Thus we are motivated to explore their computational capabilities. Here we focus on the GRU, described by the equations below:
\begin{align*}
z_t &= \sigma(W_zx_t + U_zh_{t - 1} + b_z)\\
r_t &= \sigma(W_rx_t + U_rh_{t - 1} + b_r)\\
h_t &= z_t \circ h_{t - 1} + (1 - z_t) \circ \text{tanh}(W_hx_t + U_h(r_t \circ h_{t - 1}) + b_h)\\
o_t &= f(h_t)
\end{align*}
for some $f: \mathbb{R}^{m \times 1} \rightarrow \mathbb{R}$ where $h$ has dimension $m \times 1$ and $\sigma(x) = (1 + e^{-x})^{-1}$ is the {\it{sigmoid}} function and $\text{tanh}(x) = (e^{2x} - 1)(e^{2x} + 1)^{-1}$ is the {\it{hyperbolic tangent}} function, and the $\circ$ symbol represents element-wise multiplication. Usually the hidden state $h_0$ is initialized to be $0$, but we will ignore that restriction. Some literature switches the placements of the $z_t$ and $1 - z_t$, but since $\sigma(-x) = 1 - \sigma(x)$ this is immaterial. 

\bigskip
\noindent
We begin this section by again limiting our architecture to use finite precision, and also assume $f(h) = W_oh$ for some $W_o \in \mathbb{R}^{1 \times m}$. We can prove an analogue of Theorem 1.1:

\bigskip
\noindent
{\bf{Theorem 2.1}}. For every language $L \subseteq \Sigma^*$, $L$ is regular if and only if $L$ is the $S$-language of some finite precision GRU. 

\bigskip
\noindent
{\it{Proof}}. The ``if" direction can be shown in the same manner as in Theorem 1.1. So, here we focus on the ``only if" direction. Suppose we have a DFA $D = (Q, \Sigma, \delta, q_0, F)$ with corresponding language $L$. We will construct a GRU whose inputs are one-hotted symbols from $\Sigma$ with $|Q||\Sigma|$ hidden nodes whose $\{0\}$-language is $L$.

\bigskip
\noindent
For convenience, for all $x \in \Sigma$ let $e_x$ denote the corresponding one-hot vector for $x$. Furthermore, let $N = |\Sigma||Q|$. 

\bigskip
\noindent
First set $W_z = W_h = 0$ and $U_z = U_r = 0$ and $b_z = b_r = b_h = 0$, so the simplified GRU is given by:
\begin{align*}
r_t &= \sigma(W_rx_t)\\
\text{tanh}^{-1}(2h_t - h_{t - 1}) &= U_h(r_t \circ h_{t - 1})\\
o_t &= W_oh_t
\end{align*}
Now, define an arbitrary bijective map $g: \{1, 2, \dots, |Q|\} \rightarrow Q$. Then construct $|Q|$ vectors
$$ s_i = [s_{i,1}\ \ s_{i,2}\ \ \dots\ \ s_{i,N}]^T $$
where for all $i \in \{1, 2, \dots, |Q|\}$ and $k \in \{1, 2, \dots, N\}$ we set
$$ s_{i, k} = \begin{cases}0.25, & 0 \le i|\Sigma| - k < |\Sigma|\\0, & \text{otherwise}\end{cases} $$
Our goal will be to find $W_r$ and $U_h$ such that if $h_{t - 1} = s_i$ for some $i$, and $x_t$ is the one-hot encoding of some $x \in \Sigma$, then $h_t = s_j$ where if $g(i) = q$ for some $q \in Q$ then $g(j) = \delta(q, x)$. If this is possible, then we could set $h_0 = s_{g^{-1}(q_0)}$ and be able to track the current state of the DFA effectively.

\bigskip
\noindent
The strategy for accomplishing this is essentially to pick a simple $W_r$, and then solve a system of equations to produce the desired $U_h$. 

\bigskip
\noindent
For convenience, define the natural map $h: \{1, 2, \dots, |\Sigma|\} \rightarrow \Sigma$ where $h(i) = x$ if and only if the $i$th element of $e_x$ is equal to $1$. 

\bigskip
\noindent
Let
$$ W_r = \begin{bmatrix}\sigma^{-1}(r_{1,1}) & \sigma^{-1}(r_{1,2}) & \dots & \sigma^{-1}(r_{1,|\Sigma|})\\\sigma^{-1}(r_{2,1}) & \sigma^{-1}(r_{2,2}) & \dots & \sigma^{-1}(r_{2,|\Sigma|})\\\vdots & \vdots & \ddots & \vdots\\\sigma^{-1}(r_{N,1}) & \sigma^{-1}(r_{N,2}) & \dots & \sigma^{-1}(r_{N,|\Sigma|})\end{bmatrix} $$
\\
where 
$$ r_{k, j} = \begin{cases}0.4, & k \not\equiv j \pmod{|\Sigma|}\\0.8, & k \equiv j \pmod{|\Sigma|}\end{cases} $$ 
for all $k \in \{1, 2, \dots, N\}$ and $j \in \{1, 2, \dots, |\Sigma|\}$. Now consider the $N$ equations 
$$ \text{tanh}^{-1}(2s_j - s_i) = U_h(\sigma(W_re_x) \circ s_i) $$
where $g(j) = \delta(g(i), x)$, for every $i \in \{1, 2, \dots, |Q|\}$ and $x \in \Sigma$. Let
\begin{align*}
b_{k, (i -1)|\Sigma| + j} &= \text{tanh}^{-1}(2s_{g^{-1}(\delta(g(i), h(j))), k} - s_{i, k})\\
c_{k, (i - 1)|\Sigma| + j} &= r_{k, j}s_{i, k}
\end{align*}
for all $i \in \{1, 2, \dots, |Q|\}$ and $j \in \{1, 2, \dots, |\Sigma|\}$ and $k \in \{1, 2, \dots, N\}$. Letting
\\
\begin{align*}
B = \begin{bmatrix}b_{1,1} & b_{1,2} & \dots & b_{1,N}\\b_{2,1} & b_{2,2} & \dots & b_{2,N}\\\vdots & \vdots & \ddots & \vdots\\b_{N,1} & b_{N,2} & \dots & b_{N,N}\end{bmatrix}\ \ \ \ \ \ \ \ \ 
C = \begin{bmatrix}c_{1,1} & c_{1,2} & \dots & c_{1,N}\\c_{2,1} & c_{2,2} & \dots & c_{2,N}\\\vdots & \vdots & \ddots & \vdots\\c_{N,1} & c_{N,2} & \dots & c_{N,N}\end{bmatrix}
\end{align*}
\\
The $N$ earlier equations can now be combined as a single matrix equation given by
$$ U_hC = B \Longrightarrow U_h = BC^{-1} $$

\bigskip
\noindent
Now it is easy to see that
$$ C = \begin{bmatrix}C_1 & 0 & \dots & 0\\0 & C_2 & \dots & 0\\\vdots & \vdots & \ddots & \vdots\\0 & 0 & \dots & C_{|Q|}\end{bmatrix} $$
where $C_j$ is a $|\Sigma| \times |\Sigma|$ matrix for each $j \in \{1, 2, \dots, |\Sigma|\}$. In particular, we have that
$$ C_j = \begin{bmatrix}0.2 & 0.1 & \dots & 0.1\\0.1 & 0.2 & \dots & 0.1\\\vdots & \vdots & \ddots & \vdots\\0.1 & 0.1 & \dots & 0.2\end{bmatrix} $$
for each $j$.

\bigskip
\noindent
Using basic row operations it is easy to see that $\text{det}(C_j) = 0.1^{|\Sigma|}(|\Sigma| + 1)$ for all $j$, so 
$$ \text{det}(C) = \prod_{j= 1}^{|Q|}\text{det}(C_j) = 0.1^N(|\Sigma| + 1)^{|Q|} \ne 0 $$
and thus $C^{-1}$ is well-defined. Furthermore, since $s_{i, k} \in \{0, 0.25\}$ for each $i, k$, the inputs into all inverse hyperbolic tangents in $B$ lie in $(-1, 1)$ and so $B$ is well-defined as well. Thus our expression for $U_h$ is well-defined.

\bigskip
\noindent
Now, given our choices for the $s_i, W_r$, and $U_h$, after reading any input $w = w_1w_2\dots w_m$, if $q$ is the current state of the DFA associated with $L$, then $h_m = s_{g^{-1}(q)}$. Now because the $s_i$ are clearly linearly independent, we can find a $W_o$ such that
$$ W_os_i = \begin{cases}0, & g(i) \in F\\1, & g(i) \not\in F\end{cases} $$
for all $i \in \{1, 2, \dots, Q\}$ and it's clear that the $\{0\}$-language of the resulting GRU will be $L$, as desired.\qed

\bigskip
\noindent
{\bf{Discussion 2.2}}. In the above proof, we are implicitly assuming that the activation functions of the GRU are not actually the sigmoid and hyperbolic tangent functions but rather finite precision analogues for which the equations we solved are all consistent. However, for the remainder of this section we can drop this assumption.

\bigskip
\noindent
If we remove the finite precision restriction, we again wish to prove that Gated RNNs are as powerful as PDAs. To do so, we emulate the approach from Section 1. Immediately we encounter difficulties - in particular, our previous approach relied on maintaining the digits of a state $s$ in base $2n + 1$ very carefully. With outputs now run through sigmoid and hyperbolic tangent functions, this becomes very hard. Furthermore, updating the state $s$ occasionally requires multiplication by $2n + 1$ (when we read a closing parenthesis). But because $\sigma(x) \in (0, 1)$ and $\text{tanh}(x) \in (-1, 1)$ for all $x \in \mathbb{R}$, this is impossible to do with the GRU architecture.

\bigskip
\noindent
To account for both of these issues, instead of keeping track of the state $s_t$ as we read a word, we will instead keep track of the state $s'_t$ of a word $w = w_1w_2\dots w_m \in \Sigma^*$ defined by
\begin{align*}
s'_0 &= 0\\
s'_t &= \begin{cases}(2n + 1)^{-1 - k}s'_{t - 1} + 2i(2n + 1)^{-1 - kt}, & w_t =(_i \\(2n + 1)^{1 - k}s'_{t - 1} - 2i(2n + 1)^{-kt}, & w_t =\ )_i\end{cases}
\end{align*}
for all $i \in \{1, 2, \dots, n\}$, for some predetermined sufficiently large $k$. We have the following relationship between $s'_t$ and $s_t$:

\bigskip
\noindent
{\bf{Lemma 2.3}}. For any word $w = w_1w_2\dots w_m \in \Sigma^*$ we have $s_t = (2n + 1)^{kt}s'_t$ for all $t \in \{1, 2, \dots, m\}$. 

\bigskip
\noindent
{\it{Proof}}. Multiplying the recurrence relationship for $s'_t$ by $(2n + 1)^{kt}$ we recover the recurrence relationship for $s_t$ in Section 1, implying the desired result.\qed

\bigskip
\noindent
Thus the state $s'$ allows us to keep track of the old state $s$ without having to multiply by any constant greater than $1$. Furthermore, for large $k$, $s'$ will be extremely small, allowing us to abuse the fact that $\text{tanh}(x) \sim x$ for small values of $x$. In terms of the stack of digits interpretation of $s$, $s'$ is the same except between every pop or push we add $k$ zeros to the top of the stack. 

\bigskip
\noindent
Again we wish to construct a GRU from whose hidden state we can recover $s'_t$. Ignoring the output layer for now, consider the GRU defined by
\begin{align*}
h_0 &= [h_{1, 0}\ \ 1\ \ 1]^T\\
z_t &= \sigma(W_zx_t)\\
r_t &= \sigma(W_rx_t)\\
h_t &= z_t \circ h_{t - 1} + (1 - z_t) \circ \text{tanh}(U_h(r_t \circ h_{t - 1}))
\end{align*}
where $h_{1, 0} \ge 0$ will be determined later, the inputs $x$ are again $2n \times 1$ one-hots of the symbols in $\Sigma$ in the order $(_1, (_2, \dots, (_n, )_1, )_2, \dots, )_n$ and the hidden states have dimension $3 \times 1$ where

\begin{align*}
W_z &= \begin{bmatrix}\sigma^{-1}((2n + 1)^{-1 - k}) & \dots & \sigma^{-1}((2n + 1)^{-1 - k}) & \sigma^{-1}((2n + 1)^{1 - k}) & \dots & \sigma^{-1}((2n + 1)^{1 - k})\\\sigma^{-1}((2n + 1)^{-k}) & \dots & \dots & \dots & \dots & \sigma^{-1}((2n + 1)^{-k})\\\sigma^{-1}((2n + 1)^{-k}) & \dots & \dots & \dots & \dots & \sigma^{-1}((2n + 1)^{-k})\end{bmatrix}\\
\\
W_r &= \begin{bmatrix}0 & 0 & \sigma^{-1}(0.5 - 2((2n + 1)^{k + 1} - 1)^{-1})\\0 & 0 & \sigma^{-1}(0.5 - 4((2n + 1)^{k + 1} - 1)^{-1})\\\vdots & \vdots & \vdots\\0 & 0 & \sigma^{-1}(0.5 - 2n((2n + 1)^{k + 1} - 1)^{-1})\\0 & 0 & \sigma^{-1}(0.5 + 2((2n + 1)^{k} - 2n - 1)^{-1})\\0 & 0 & \sigma^{-1}(0.5 + 4((2n + 1)^{k} - 2n - 1)^{-1})\\\vdots & \vdots & \vdots\\0 & 0 & \sigma^{-1}(0.5 + 2n((2n + 1)^{k} - 2n - 1)^{-1})\end{bmatrix}^T\\
\\
U_h &= \begin{bmatrix}0 & 1 & -1\\0 & 0 & 0\\0 & 0 & 0\end{bmatrix}\\
\end{align*}
where $\sigma^{-1}(x) = -\ln(x^{-1} - 1)$ is the inverse of the sigmoid function. For sufficiently large $k$, clearly our use of $\sigma^{-1}$ is well-defined. We will show the following invariant:

\bigskip
\noindent
{\bf{Lemma 2.4}}. Given an input word $w \in P_n \cup D_n$, if $h_{1, 0} = 0$ then we have $h_t \approx [s'_t\ \ (2n + 1)^{-kt}\ \ (2n + 1)^{-kt}]^T$ for all $t$.

\bigskip
\noindent
{\it{Proof}}. As in Section 1, let $z_t = [z_{1, t}\ \ z_{2, t}\ \ z_{3, t}]^T$ and $r_t = [r_{1, t}\ \ r_{2, t}\ \ r_{3, t}]^T$ and $h_t = [h_{1, t}\ \ h_{2, t}\ \ h_{3, t}]^T$. First, we will show $h_{2, t} = (2n + 1)^{-kt}$ for all $t \in \{1, 2, \dots, m\}$ by induction on $t$. The base case is trivial, so note
\begin{align*}
z_{2, t + 1} &= \sigma(\sigma^{-1}((2n + 1)^{-k})) = (2n + 1)^{-k}\\
r_{2, t + 1} &= \sigma(0) = 0.5\\
h_{2, t + 1} &= z_{2, t + 1}h_{2, t} + (1 - z_{2, t + 1})\text{tanh}(0) = (2n + 1)^{-k}h_{2, t}
\end{align*}
so by induction $h_{2, t + 1} = (2n + 1)^{-k(t + 1)}$ as desired. Similarly, we obtain $h_{3, t} = (2n + 1)^{-kt}$ for all $t$. 

\bigskip
\noindent
Now we restrict our attention to $h_{1, t}$. Note that
\begin{align*}
z_{1, t} &= \begin{cases}\sigma(\sigma^{-1}((2n + 1)^{-1 - k})) = (2n + 1)^{-1 - k}, & w_t = (_i\\\sigma(\sigma^{-1}((2n + 1)^{1 - k})) = (2n + 1)^{1 - k}, & w_t =\ )_i\end{cases}\\
r_{2, t} &= \sigma(0) = 0.5\\
r_{3, t} &= \begin{cases}0.5 - 2i((2n + 1)^{k + 1} - 1)^{-1}, & w_t = (_i\\0.5 + 2i((2n + 1)^{k} - 2n - 1)^{-1}, & w_t =\ )_i\end{cases}\\
\end{align*}
and so using the definition of $U_h$ we obtain
\begin{align*}
h_{1, t} &= z_{1, t}h_{1, t - 1} + (1 - z_{1, t})\text{tanh}(2^{-k(t - 1)}r_{2, t} - 2^{-k(t - 1)}r_{3, t})\\
&= \begin{cases}(2n + 1)^{-1 - k}h_{1, t - 1} + (1 - (2n + 1)^{-1 - k})\text{tanh}(2i(2n + 1)^{-k(t - 1)}((2n + 1)^{k + 1} - 1)^{-1}), & w_t = (_i\\(2n + 1)^{1 - k}h_{1, t - 1} - (1 - (2n + 1)^{1 - k})\text{tanh}(2i(2n + 1)^{-k(t - 1)}((2n + 1)^{k} - 2n - 1)^{-1}), & w_t =\ )_i\end{cases}
\end{align*}
If we removed the $\text{tanh}$ from the above expression, it would simplify to 
$$ h_{1, t} = \begin{cases}(2n + 1)^{-1 - k}h_{1, t - 1} + 2i(2n + 1)^{-1 -kt}, & w_t = (_i\\(2n + 1)^{1 - k}h_{1, t - 1} - 2i(2n + 1)^{-kt}, & w_t =\ )_i\end{cases} $$
which is exactly the recurrence relation satisfied by $s'_t$. Since the expressions inside the hyperbolic tangents are extremely small (on the order of $2^{-kt}$), this implies that $h_{1, t}$ is a good approximation for $s'_t$ as desired. This will be formalized in the next lemma.\qed

\bigskip
\noindent
{\bf{Lemma 2.5}}. For any input word $w \in P_n \cup D_n$, if $h_{1, 0} = 0$ then we have $|(2n + 1)^{kt}h_{1, t} - s_t| < 2(2n + 1)^{-2k + 7}$ for all $t$. 

\bigskip
\noindent
{\it{Proof}}. Let $\epsilon_t = (2n + 1)^{kt}h_{1, t} - s_t$ for all $t$. Then we easily find that
$$ \epsilon_t = \begin{cases}(2n + 1)^{-1}\epsilon_{t - 1} + (2n + 1)^{kt}(1 - (2n + 1)^{-1-k})\text{tanh}\left(\frac{2i(2n + 1)^{-k(t - 1)}}{(2n + 1)^{k + 1} - 1}\right) - \frac{2i}{2n + 1}, & w_t = (_i\\(2n + 1)\epsilon_{t - 1} - (2n + 1)^{kt}(1 - (2n + 1)^{1-k})\text{tanh}\left(\frac{2i(2n + 1)^{-k(t - 1)}}{(2n + 1)^{k} - 2n - 1}\right) + 2i, & w_t = \ )_i\end{cases} $$
Now define $\epsilon'_t$ by the recurrence
$$ \epsilon'_t = \begin{cases}(2n + 1)^{-1}\epsilon'_{t - 1} - (2n + 1)^{kt}(1 - (2n + 1)^{-1-k})\text{tanh}\left(\frac{2i(2n + 1)^{-k(t - 1)}}{(2n + 1)^{k + 1} - 1}\right) + \frac{2i}{2n + 1}, & w_t = (_i\\(2n + 1)\epsilon'_{t - 1} - (2n + 1)^{kt}(1 - (2n + 1)^{1-k})\text{tanh}\left(\frac{2i(2n + 1)^{-k(t - 1)}}{(2n + 1)^{k} - 2n - 1}\right) + 2i, & w_t = \ )_i\end{cases} $$
with $\epsilon'_0 = \epsilon_0 = 0$. Because $\text{tanh}(x) < x$ for all $x > 0$ it is easy to see that $\epsilon'_t \ge |\epsilon_t|$ for all $t$. 

\bigskip
\noindent
Now by a Taylor expansion, $\text{tanh}(x) = x - \frac{x^3}{3} + \frac{2x^5}{15} + O(x^7)$, so we have that
$$ 0 \le x - \text{tanh}(x) \le \frac{x^3}{3} < x^3 $$
for $x > 0$. Thus we obtain the bound
$$ \frac{2i}{2n + 1} - (2n + 1)^{kt}(1 - (2n + 1)^{-1-k})\text{tanh}\left(\frac{2i(2n + 1)^{-k(t - 1)}}{(2n + 1)^{k + 1} - 1}\right) < \frac{8i^3(2n + 1)^{-2kt + 2k - 1}}{((2n + 1)^{k + 1} - 1)^2} $$
Since $2i < 2n + 1$ and $(2n + 1)^{k + 1} - 1 \ge (2n + 1)^k$ we also have 
$$ \frac{8i^3(2n + 1)^{-2kt + 2k - 1}}{((2n + 1)^{k + 1} - 1)^2} < \frac{(2n + 1)^3(2n + 1)^{-2kt + 2k - 1}}{(2n + 1)^{2k}} = (2n + 1)^{-2kt + 2} < (2n + 1)^{-2kt + 5} $$
Similarly we obtain the bound 
$$ 2i - (2n + 1)^{kt}(1 - (2n + 1)^{1-k})\text{tanh}\left(\frac{2i(2n + 1)^{-k(t - 1)}}{(2n + 1)^{k} - 2n - 1}\right) < \frac{8i^3(2n + 1)^{-2kt + 2k - 2}}{((2n + 1)^{k - 1} - 1)^2} $$
Since again $2i < 2n + 1$ and $(2n + 1)^{k - 1} - 1 \ge (2n + 1)^{k - 2}$ we also have
$$ \frac{8i^3(2n + 1)^{-2kt + 2k - 2}}{((2n + 1)^{k - 1} - 1)^2} < \frac{(2n + 1)^3(2n + 1)^{-2kt + 2k - 2}}{(2n + 1)^{2k - 4}} = (2n + 1)^{-2kt + 5} $$
Thus if we define $a_t$ by the recurrence
$$ a_t = \begin{cases}(2n + 1)^{-1}a_{t - 1} + (2n + 1)^{-2kt + 5}, & w_t = (_i\\(2n + 1)a_{t - 1} + (2n + 1)^{-2kt + 5}, & w_t = \ )_i\end{cases} $$
with $a_0 = \epsilon'_0 = 0$, then $a_t \ge \epsilon'_t$ for all $t$. 

\bigskip
\noindent
Now we wish to upper bound $a_t$. Since $i$ is not present in the recurrence for $a_t$, assume without loss of generality that all parenthesis in an input word $w = w_1w_2\dots w_m \in P_n \cup D_n$ lie in $\{(_1, )_1\}$. Suppose that $)_1(_1$ was a substring of $w$, so that $w = x)_1(_1y$. Then we would have 
\begin{align*}
a_{|x| + 2} &= (2n + 1)^{-1}\left((2n + 1)a_{|x|} + (2n + 1)^{-2k(|x| + 1) + 5}\right) + (2n + 1)^{-2k(|x| + 2) + 5}\\
&= a_{|x|} + (2n + 1)^{-2k(|x| + 1) + 4} + (2n + 1)^{-2k(|x| + 2) + 5}
\end{align*}
However, for the word $w' = x(_1)_1y$ (which would clearly still lie in $P_n \cup D_n$) we would have
\begin{align*}
a_{|x| + 2} &= (2n + 1)\left((2n + 1)^{-1}a_{|x|} + (2n + 1)^{-2k(|x| + 1) + 5}\right) + (2n + 1)^{-2k(|x| + 2) + 5}\\
&= a_{|x|} + (2n + 1)^{-2k(|x| + 1) + 6} + (2n + 1)^{-2k(|x| + 2) + 5}
\end{align*}
which is larger. Thus to upper bound $a_t$ it suffices to consider only words that do not contain the substring $)_1(_1$, which are words in the form
$$ w = (_1(_1\dots(_1)_1)_1\dots)_1 $$
with $r$ open parentheses followed by $s \le r$ closing parentheses. Furthermore, adding extra closing parenthesis where suitable clearly increases the final $a_t$ so we can assume $s = r$. We can then exactly calculate $a_{2r}$ as 
$$ \sum_{i = 1}^{r}(2n + 1)^{-2ki + 5 + i} + \sum_{i = 1}^{r}(2n + 1)^{-2k(r + i) + 5 + r - i} $$
Considering each sum separately we have for sufficiently large $k$ that
\begin{align*}
\sum_{i = 1}^{r}(2n + 1)^{-2ki + 5 + i} &< \lim_{q \rightarrow \infty}(2n + 1)^5\sum_{i = 1}^{q}(2n + 1)^{(-2k + 1)i}\\
&= \frac{(2n + 1)^{-2k + 6}}{1 - (2n + 1)^{-2k + 1}}\\
&< (2n + 1)^{-2k + 7}
\end{align*}
and 
\begin{align*}
\sum_{i = 1}^{r}(2n + 1)^{-2k(r + i) + 5 + r - i} &= (2n + 1)^{5 + (1 - 2k)r}\sum_{i = 1}^{r}(2n + 1)^{(-2k - 1)i}\\
&< \frac{(2n + 1)^{5 + (1 - 2k)r}}{1 - (2n + 1)^{-2k - 1}}\\
&< (2n + 1)^{-2k + 7}
\end{align*}
And therefore $2(2n + 1)^{-2k + 7}$ is an upper bound on $a_t$. Thus
$$ |\epsilon_t| \le \epsilon'_t \le a_t < 2(2n + 1)^{-2k + 7} $$
for all $t$ as desired.\qed

\bigskip
\noindent
{\bf{Corollary 2.6}}. For any input word $w = w_1w_2\dots w_m \in P_n \cup D_n$, if $w_1w_2\dots w_t$ contains $a_t$ open parentheses and $b_t \le a_t$ closing parentheses then 
$$ (2n + 1)^{kt}h_{1, t} = h_{1, 0}(2n + 1)^{b_t - a_t} + s_t + \epsilon $$
with $|\epsilon| < 2(2n + 1)^{-2k + 7}$ for all $t$. 

\bigskip
\noindent
{\it{Proof}}. This follows directly from the computations in the proof of Lemma 2.5 and the recurrence for $h_{1, t}$.\qed

\bigskip
\noindent
Now, set $h_{1, 0} = 3(2n + 1)^{-2k + 7}$. We then have the following useful analogues of Lemmas 1.7 and 1.8:

\bigskip
\noindent
{\bf{Corollary 2.7}}. For any input word $w = w_1w_2\dots w_m \in P_n \cup D_n$ we have $h_{1, m} > 0$.

\bigskip
\noindent
{\it{Proof}}. This follows immediately from Corollary 2.6 and the fact that $h_{1, 0} > 2(2n + 1)^{-2k + 7}$. \qed

\bigskip
\noindent
{\bf{Lemma 2.8}}. Given a word $w_1w_2\dots w_m \in P_n$, consider the unique $j \in \{1, 2, \dots, n\}$ such that $w)_j \in P_n \cup D_n$. Then for an input word $w)_i$ with $i > j$, we have $h_{1, m + 1} < 0$. 

\bigskip
\noindent
Note that
$$ h_{1, m + 1} = (2n + 1)^{1 - k}h_{1, m} - (1 - (2n + 1)^{1 - k})\text{tanh}(2i(2n + 1)^{-km}((2n + 1)^{k} - 2n - 1)^{-1}) $$
so multiplying both sides by $(2n + 1)^{k(m + 1)}$ and using the inequality from the proof of Lemma 2.5 we have
$$ (2n + 1)^{k(m + 1)}h_{1, m + 1} < (2n + 1)^{km + 1}h_{1, m} - 2i + (2n + 1)^{-2k(m + 1) + 5} $$
Now by Corollary 2.6 we have that
$$ (2n + 1)^{km}h_{1, m} < s_m + 5(2n + 1)^{-2k + 7} < \frac{2j + 1}{2n + 1} + 5(2n + 1)^{-2k + 7} $$
where we used the inequality from the proof of Lemma 1.6 and the fact that $h_{1, 0} = 3(2n + 1)^{-2k + 7}$. Therefore
$$ (2n + 1)^{k(m + 1)}h_{1, m + 1} < 2j + 1 - 2i + 5(2n + 1)^{-2k + 8} + (2n + 1)^{-2k(m + 1) + 5} $$
Since $i > j$ we have that $2j + 1 - 2i \le -1$ and so for sufficiently large $k$ we then have
$$ h_{1, m + 1} < 0 $$
as desired. \qed

\bigskip
\noindent
With these results in hand, consider the larger GRU, an example of which can be found in Appendix A.2, defined by
\begin{align*}
h_0 &= [3(2n + 1)^{-2k + 7}\ \ 1\ \ 1\ \ 1\ \ 3(2n + 1)^{-2k + 7}\ \ 1\ \ 1\ \ 1]^T\\
z_t &= \sigma(W_zx_t + U_zh_{t - 1})\\
r_t &= \sigma(W_rx_t)\\
h_t &= z_t \circ h_{t - 1} + (1 - z_t) \circ \text{tanh}(U_h(r_t \circ h_{t - 1}))\\
o_t &= \frac{|h_{1, t}|}{h_{2, t}} - h_{4, t} - h_{8, t} + 2
\end{align*}
where the inputs $x$ are again $2n \times 1$ one-hots of the symbols in $\Sigma$ in the order $(_1, (_2, \dots, (_n, )_1, )_2, \dots, )_n$ and the hidden states have dimension $8 \times 1$ where

\begin{align*}
W_z &= \begin{bmatrix}\sigma^{-1}((2n + 1)^{-1 - k}) & \dots & \sigma^{-1}((2n + 1)^{-1 - k}) & \sigma^{-1}((2n + 1)^{1 - k}) & \dots & \sigma^{-1}((2n + 1)^{1 - k})\\\sigma^{-1}((2n + 1)^{-k}) & \dots & \dots & \dots & \dots & \sigma^{-1}((2n + 1)^{-k})\\\sigma^{-1}((2n + 1)^{-k}) & \dots & \dots & \dots & \dots & \sigma^{-1}((2n + 1)^{-k})\\0 & \dots & \dots & \dots & \dots & 0\\\sigma^{-1}((2n + 1)^{-1 - k}) & \dots & \sigma^{-1}((2n + 1)^{-1 - k}) & \sigma^{-1}((2n + 1)^{1 - k}) & \dots & \sigma^{-1}((2n + 1)^{1 - k})\\\sigma^{-1}((2n + 1)^{-k}) & \dots & \dots & \dots & \dots & \sigma^{-1}((2n + 1)^{-k})\\\sigma^{-1}((2n + 1)^{-k}) & \dots & \dots & \dots & \dots & \sigma^{-1}((2n + 1)^{-k})\\0 & \dots & \dots & \dots & \dots & 0\end{bmatrix}\\
\\
U_z &= \begin{bmatrix}0 & 0 & 0 & 0 & 0 & 0 & 0 & 0\\0 & 0 & 0 & 0 & 0 & 0 & 0 & 0\\0 & 0 & 0 & 0 & 0 & 0 & 0 & 0\\\infty & 0 & 0 & 0 & 0 & 0 & 0 & 0\\0 & 0 & 0 & 0 & 0 & 0 & 0 & 0\\0 & 0 & 0 & 0 & 0 & 0 & 0 & 0\\0 & 0 & 0 & 0 & 0 & 0 & 0 & 0\\0 & 0 & 0 & 0 & \infty & 0 & 0 & 0\end{bmatrix}\\
\\
W_r &= \begin{bmatrix}0 & 0 & \sigma^{-1}(0.5 - 2((2n + 1)^{k + 1} - 1)^{-1}) & 0 & 0 & 0 & \sigma^{-1}(0.5 - 2n((2n + 1)^{k + 1} - 1)^{-1}) & 0\\0 & 0 & \sigma^{-1}(0.5 - 4((2n + 1)^{k + 1} - 1)^{-1}) & 0 & 0 & 0 & \sigma^{-1}(0.5 - (2n - 2)((2n + 1)^{k + 1} - 1)^{-1}) & 0\\\vdots & \vdots & \vdots & \vdots & \vdots & \vdots & \vdots & \vdots\\0 & 0 & \sigma^{-1}(0.5 - 2n((2n + 1)^{k + 1} - 1)^{-1}) & 0 & 0 & 0 & \sigma^{-1}(0.5 - 2((2n + 1)^{k + 1} - 1)^{-1}) & 0\\0 & 0 & \sigma^{-1}(0.5 + 2((2n + 1)^{k} - 2n - 1)^{-1}) & 0 & 0 & 0 & \sigma^{-1}(0.5 + 2n((2n + 1)^{k} - 2n - 1)^{-1}) & 0\\0 & 0 & \sigma^{-1}(0.5 + 4((2n + 1)^{k} - 2n - 1)^{-1}) & 0 & 0 & 0 & \sigma^{-1}(0.5 + (2n - 2)((2n + 1)^{k} - 2n - 1)^{-1}) & 0\\\vdots & \vdots & \vdots & \vdots & \vdots & \vdots & \vdots & \vdots\\0 & 0 & \sigma^{-1}(0.5 + 2n((2n + 1)^{k} - 2n - 1)^{-1}) & 0 & 0 & 0 & \sigma^{-1}(0.5 + 2((2n + 1)^{k} - 2n - 1)^{-1}) & 0\end{bmatrix}^T\\
\\
U_h &= \begin{bmatrix}0 & 1 & -1 & 0 & 0 & 0 & 0 & 0\\0 & 0 & 0 & 0 & 0 & 0 & 0 & 0\\0 & 0 & 0 & 0 & 0 & 0 & 0 & 0\\0 & 0 & 0 & 0 & 0 & 0 & 0 & 0\\0 & 0 & 0 & 0 & 0 & 1 & -1 & 0\\0 & 0 & 0 & 0 & 0 & 0 & 0 & 0\\0 & 0 & 0 & 0 & 0 & 0 & 0 & 0\\0 & 0 & 0 & 0 & 0 & 0 & 0 & 0\end{bmatrix}\\
\end{align*}
As before, with respect to a word $w \in \Sigma^*$ define $s_t$ by
\begin{align*}
s_0 &= 0\\
s_t &= \begin{cases}(2n + 1)^{-1}s_{t - 1} + 2i(2n + 1)^{-1}, & w_t =(_i \\(2n + 1)s_{t - 1} - 2i, & w_t =\ )_i\end{cases}
\end{align*}
for all $i \in \{1, 2, \dots, n\}$ and all $t$. Similarly define $\overline{s}_t$ by
\begin{align*}
\overline{s}_0 &= 0\\
\overline{s}_t &= \begin{cases}(2n + 1)^{-1}\overline{s}_{t - 1} + 2i(2n + 1)^{-1}, & w_t =(_{n - i} \\(2n + 1)\overline{s}_{t - 1} - 2i, & w_t =\ )_{n - i}\end{cases}
\end{align*}
For our new GRU, let $h_t = [h_{1, t}\ \ h_{2, t}\ \ h_{3, t}\ \ h_{4, t}\ \ h_{5, t}\ \ h_{6, t}\ \ h_{7, t}\ \ h_{8, t}]^T$. We then have the following results:

\bigskip
\noindent
{\bf{Lemma 2.9}}. For any input word $w \in \Sigma^*$ we have $h_{2, t} = h_{3, t} = h_{6, t} = h_{7, t} = (2n + 1)^{-kt}$.

\bigskip
\noindent
{\it{Proof}}. This follows immediately from the proof of Lemma 2.4.\qed

\bigskip
\noindent
{\bf{Lemma 2.10}}. For any input word $w = w_1w_2\dots w_m \in P_n \cup D_n$, if $w_1w_2\dots w_t$ contains $a_t$ open parentheses and $b_t \le a_t$ closing parenthesis then 
$$ (2n + 1)^{kt}h_{1, t} = h_{1, 0}(2n + 1)^{b_t - a_t} + s_t + \epsilon_1 $$
$$ (2n + 1)^{kt}h_{5, t} = h_{5, 0}(2n + 1)^{b_t - a_t} + \overline{s}_t + \epsilon_2 $$
with $|\epsilon_1|, |\epsilon_2| < 2(2n + 1)^{-2k + 7}$ for all $t$. 

\bigskip
\noindent
{\it{Proof}}. This follows immediately from the proof of Corollary 2.6 and the new $W_r$, since $h_{5, t}$ behaves exactly like $h_{1, t}$ if each input $(_i$ or $)_i$ were $(_{n - i}$ or $)_{n - i}$ respectively, instead. \qed

\bigskip
\noindent
{\bf{Lemma 2.11}}. For any input word $w = w_1w_2\dots w_m \in \Sigma^*$ we have $h_{4, m}, h_{8, m} \in \{0, 1\}$ and $h_{4, m} = h_{8, m} = 1 $ if and only if $w_1w_2\dots w_{m - 1} \in P_n \cup D_n$.

\bigskip
\noindent
{\it{Proof}}. From our chosen $U_z$ we see that 
$$ z_{4, t} = \begin{cases}\sigma(\infty) = 1, & h_{1, t-1} > 0\\\sigma(-\infty) = 0, & h_{1, t - 1} < 0\end{cases} $$
$$ z_{8, t} = \begin{cases}\sigma(\infty) = 1, & h_{5, t-1} > 0\\\sigma(-\infty) = 0, & h_{5, t - 1} < 0\end{cases} $$
Since $h_{4, 0} = h_{8, 0} = 1$ and since the fourth and eighth rows of $U_h$ are identically $0$, the equation
$$ h_t = z_t \circ h_{t - 1} + (1 - z_t) \circ \text{tanh}(U_h(r_t \circ h_{t - 1})) $$
implies that
$$ h_{4, m} = \prod_{i = 1}^{m}z_{4, i} $$
$$ h_{8, m} = \prod_{i = 1}^{m}z_{8, i} $$
which immediately implies that $h_{4, m}, h_{8, m} \in \{0, 1\}$. Now, suppose $w_1w_2\dots w_{m - 1} \in P_n \cup D_n$. Then from Corollary 2.7 and its analogue for $h_{5, t}$ we see that $z_{4, t} = z_{8, t} = 1$ for all $t \in \{1, 2, \dots, m\}$, so $h_{4, m} = h_{8, m} = 1$ as desired. 

\bigskip
\noindent
Otherwise, there exists some minimal $k \in \{0, 1, \dots, m - 2\}$ such that $w_1w_2\dots w_{k + 1} \not\in P_n \cup D_n$. Then $w_{k + 1} = )_i$ for some $i \in \{1, 2, \dots, n\}$. Consider the unique $j \ne i$ such that $w_1w_2\dots w_k)_j \in P_n \cup D_n$. If $i > j$ then from the proof of Lemma 2.8 we have that $h_{1, k + 1} < 0$ and so $z_{4, k + 2} = 0$. Since $k + 2 \le m$ this means that $h_{4, m} = 0$. If $i < j$ then from the analogue of the proof of Lemma 2.8 for $h_{5, t}$, we obtain $h_{8, m} = 0$. This completes the proof. \qed

\bigskip
\noindent
We are now ready to combine these lemmas to prove an important result, the analogue of Theorem 1.10 for GRUs:

\bigskip
\noindent
{\bf{Theorem 2.12}}. The $(0, (2n + 1)^{-1})$-language of the GRU described earlier in the section is $D_n$.

\bigskip
\noindent
{\it{Proof}}. Consider any input word $w = w_1w_2\dots w_m \in \Sigma^*$ into the GRU. We consider four cases:
\begin{itemize}
\item 
Case 1: $w \in D_n$.
\end{itemize}
In this case, we clearly have $s_m = 0$ and $h_{1, m} > 0$ from the proof of Corollary 2.7, so by Lemmas 2.9 and 2.10 we have that
$$ \frac{|h_{1, m}|}{h_{2, m}} = (2n + 1)^{km}h_{1, m} = h_{1, 0} + \epsilon $$
with $|\epsilon| < 2(2n + 1)^{-2k + 7}$. Furthermore from Lemma 2.11 we have that $h_{4, m} = h_{8, m} = 1$ so since $h_{1, 0} = 3(2n + 1)^{-2k + 7}$ we must have 
$$ o_m \in ((2n + 1)^{-2k + 7}, 5(2n + 1)^{-2k + 7}) \subset (0, (2n + 1)^{-1}) $$
for sufficiently large $k$, as desired.
\begin{itemize}
\item
Case 2: $w \in P_n$.
\end{itemize}
As in Case 1 we have that $h_{1, m} > 0$ and so by Lemmas 2.9 and 2.10 we have that
$$ \frac{|h_{1, m}|}{h_{2, m}} = (2n + 1)^{km}h_{1, m} \ge s_m + \epsilon $$
with $|\epsilon| < 2(2n + 1)^{-2k + 7}$. Furthermore from Lemma 2.11 we have that $h_{4, m} = h_{8, m} = 1$ so here
$$ o_m \ge s_m - 2(2n + 1)^{-2k + 7} \ge 2(2n + 1)^{-1} - 2(2n + 1)^{-2k + 7} > (2n + 1)^{-1} $$
for sufficiently large $k$, since the minimum value of $s_m$ is clearly $2(2n + 1)^{-1}$. 
\begin{itemize}
\item 
Case 3: $w \not\in P_n \cup D_n$ and $w_1w_2\dots w_{m - 1} \in P_n \cup D_n$.
\end{itemize}
Suppose $w_1w_2\dots w_{m - 1})_j \in P_n \cup D_n$ for some unique $j \in \{1, 2, \dots, n\}$. If $w_m = )_i$ for some $i > j$ then from Lemmas 2.9 and 2.10 and the proof of Lemma 2.8 we obtain
$$ \frac{h_{1, m}}{h_{2, m}} = (2n + 1)^{km}h_{1, m} < -1 + 5(2n + 1)^{-2k + 8} + (2n + 1)^{-2km + 5} < -(2n + 1)^{-1} $$
for sufficiently large $k$. If instead $i < j$ then the same technique with the inequality $\text{tanh}(x) < x$ can be used to show 
$$ \frac{h_{1, m}}{h_{2, m}} > (2n + 1)s_m - 2(2n + 1)^{-2k + 8} - 2i > 2 - 2(2n + 1)^{-2k + 8} > (2n + 1)^{-1} $$
if $k$ is sufficiently large. As before using Lemma 2.11 we have that $h_{4, m} = h_{8, m} = 1$ and combining these bounds we find that
$$ o_m > (2n + 1)^{-1} $$
\begin{itemize}
\item 
Case 4: $w_1w_2\dots w_k \not\in P_n \cup D_n$ for some $k \in \{1, 2, \dots, m - 1\}$
\end{itemize}
In this case we know that $h_{2, m} \ge 0$ by Lemma 2.9, so we have 
$$ \frac{|h_{1, m}|}{h_{2, m}} \ge 0 $$
and by Lemma 2.11 we know that $0 \le h_{4, m} + h_{8, m} \le 1$ so 
$$ o_m \ge 0 + 2 - h_{4, m} - h_{8, m} \ge 1 > (2n + 1)^{-1} $$

\bigskip
\noindent
Thus $o_m \in (0, (2n + 1)^{-1})$ if $w \in D_n$ and $o_m > (2n + 1)^{-1}$ otherwise, as desired.\qed

\bigskip
\noindent
We may now proceed to show the main theorem of this section, an analogue of Theorem 1.11 for GRUs:

\bigskip
\noindent
{\bf{Theorem 2.13}}. For any context-free language $L$ suppose we relabel and write $L = D_n \cap R$ for some regular language $R$, whose corresponding minimum DFA has $r$ states. Then there exists a GRU with a hidden layer of size $8 + 2nr$ whose $(0, (2n + 1)^{-1})$-language is $L$. 

\bigskip
\noindent
{\it{Proof}}. This follows by combining the GRUs from the proofs of Theorems 2.1 and 2.12, as we did for simple RNNs in the proof of Theorem 1.11.\qed

\bigskip
\noindent
{\bf{Discussion 2.14}}. A critical idea in this section was to use the fact that $\text{tanh}(x) = x + O(x^2)$ near $x = 0$, and in fact this idea can be used for any activation function with a well-behaved Taylor series expansion around $x = 0$. 

\bigskip
\noindent
{\bf{Discussion 2.15}}. We ``cheated" a little bit by allowing $\infty$ edge weights and by having $o_t = f(h_t)$ where $f$ wasn't quite linear. However, $\infty$ edge weights make sense in the context of allowing infinite precision, and simple nonlinear functions over the hidden nodes are often used in practice, like the common softmax activation function.

\section{Suggestions for Further Research}
We recognize two main avenues for further research. The first is to remove the necessity for infinite edge weights in the proof of Theorem 2.13, and the second is to extend the results of Theorems 1.11 and 2.13 to Turing recognizable languages.

\bigskip
\noindent
In the proof of Lemma 2.11, edge weights of $\infty$ are necessary for determining whether a hidden node ever becomes negative. Merely using large but finite weights does not suffice, because the values in the hidden state that they will be multiplied with are rapidly decreasing. Their product will vanish, and thus we would not be able to utilize the squashing properties of common activation functions as we did in the proof of Lemma 2.11. Currently we believe that it is possible to prove that GRUs are as computationally powerful as PDAs without using infinite edge weights, but are unaware of a method to do so.

\bigskip
\noindent
Because to the our knowledge there is no analogue of the Chomsky-Sch$\ddot{\text{u}}$tzenberger Theorem for Turing recognizable languages, it seems difficult to directly extend our methods to prove that recurrent architectures are as computationally powerful as Turing machines. However, just as PDAs can lazily be described as a DFA with an associated stack, it is well-known that Turing machines are equally as powerful as DFAs with associated queues, which can be simulated with two stacks. Such an approach using two counters was used in proofs in [6], [8] to establish that RNNs with arbitrary precision can emulate Turing machines. We believe that an approach related to this fact could ultimately prove successful, but it would be more useful if set up as in the proofs above in a way that is faithful to the architecture of the neural networks.  Counter automata of this sort are also quite unlike the usual implementations found for context-free languages or their extensions for natural languages.  Work described in [10] demonstrates that in practice, LSTMs cannot really generalize to recognize the Dyck language $D_2$.  It remains to investigate whether any recent neural network variation does in fact readily generalize outside its training set to “out of sample” examples.  This would be an additional topic for future research. 

\newpage
\section*{A.1. Simple RNN $D_2$ Examples}
Consider the RNN described in the proof of Theorem 1.10 for $n = 2$. We will show the evolution of its hidden state as it reads various inputs:
\begin{itemize}
\item
Input: $w = (_2(_1)_1(_1(_2)_2)_1)_2$
\end{itemize}
For this example we obtain
\begin{align*}
h_0 &= [0\ \ 0\ \ 0\ \ 0\ \ 0\ \ 0]^T\\
h_1 &= [0.8\ \ 0\ \ 0\ \ 0\ \ 0\ \ 0]^T\\
h_2 &= [0.56\ \ 0\ \ 0\ \ 0\ \ 0\ \ 0]^T\\
h_3 &= [0\ \ 0.8\ \ 0\ \ 0\ \ 0\ \ 0]^T\\
h_4 &= [0.56\ \ 0\ \ 0\ \ 0\ \ 0\ \ 0]^T\\
h_5 &= [0.912\ \ 0\ \ 0\ \ 0\ \ 0\ \ 0]^T\\
h_6 &= [0\ \ 0.56\ \ 0\ \ 0\ \ 0\ \ 0]^T\\
h_7 &= [0\ \ 0.8\ \ 0\ \ 0\ \ 0\ \ 0]^T\\
h_8 &= [0\ \ 0\ \ 0\ \ 0\ \ 0\ \ 0]^T\\
o_8 &= 0
\end{align*}
\begin{itemize}
\item
Input: $w = (_1)_1(_2(_1)_1$
\end{itemize}
For this example we obtain
\begin{align*}
h_0 &= [0\ \ 0\ \ 0\ \ 0\ \ 0\ \ 0]^T\\
h_1 &= [0.4\ \ 0\ \ 0\ \ 0\ \ 0\ \ 0]^T\\
h_2 &= [0\ \ 0\ \ 0\ \ 0\ \ 0\ \ 0]^T\\
h_3 &= [0.8\ \ 0\ \ 0\ \ 0\ \ 0\ \ 0]^T\\
h_4 &= [0.56\ \ 0\ \ 0\ \ 0\ \ 0\ \ 0]^T\\
h_5 &= [0\ \ 0.8\ \ 0\ \ 0\ \ 0\ \ 0]^T\\
o_5 &= 0.8
\end{align*}
\begin{itemize}
\item
Input: $w = (_2)_1(_1)_2)_2$
\end{itemize}
For this example we obtain
\begin{align*}
h_0 &= [0\ \ 0\ \ 0\ \ 0\ \ 0\ \ 0]^T\\
h_1 &= [0.8\ \ 0\ \ 0\ \ 0\ \ 0\ \ 0]^T\\
h_2 &= [0\ \ 2\ \ 0\ \ 0\ \ 1\ \ 0]^T\\
h_3 &= [0.8\ \ 5\ \ 0\ \ 0\ \ 5\ \ 1]^T\\
h_4 &= [0\ \ 25\ \ 0\ \ 0\ \ 24\ \ 6]^T\\
h_5 &= [0\ \ 121\ \ 0\ \ 0\ \ 120\ \ 30]^T\\
o_5 &= 271
\end{align*}

\newpage
\section*{A.2. GRU $D_2$ Examples}
Consider the GRU described in the proof of Theorem 2.12 for $n = 2$ and $k = 5$. We will show the evolution of its hidden state as it reads various inputs:
\begin{itemize}
\item
Input: $w = (_2(_1)_1(_1(_2)_2)_1)_2$
\end{itemize}
For this example we obtain
\begin{align*}
h_0 &= [2.4\text{e-}02\ \ 1.0\ \ 1.0\ \ 1.0\ \ 2.4\text{e-}02\ \ 1.0\ \ 1.0\ \ 1.0]^T\\
h_1 &= [2.58\text{e-}04\ \ 3.20\text{e-}04\ \ 3.20\text{e-}04\ \ 1.0\ \ 1.3\text{e-}04\ \ 3.20\text{e-}04\ \ 3.20\text{e-}04\ \ 1.0]^T\\
h_2 &= [5.74\text{e-}08\ \ 1.02\text{e-}07\ \ 1.02\text{e-}07\ \ 1.0\ \ 9.02\text{e-}08\ \ 1.02\text{e-}07\ \ 1.02\text{e-}07\ \ 1.0]^T\\
h_3 &= [2.64\text{e-}11\ \ 3.28\text{e-}11\ \ 3.28\text{e-}11\ \ 1.0\ \ 1.33\text{e-}11\ \ 3.28\text{e-}11\ \ 3.28\text{e-}11\ \ 1.0]^T\\
h_4 &= [5.88\text{e-}15\ \ 1.05\text{e-}14\ \ 1.05\text{e-}14\ \ 1.0\ \ 9.24\text{e-}15\ \ 1.05\text{e-}14\ \ 1.05\text{e-}14\ \ 1.0]^T\\
h_5 &= [3.06\text{e-}18\ \ 3.36\text{e-}18\ \ 3.36\text{e-}18\ \ 1.0\ \ 1.93\text{e-}18\ \ 3.36\text{e-}18\ \ 3.36\text{e-}18\ \ 1.0]^T\\
h_6 &= [6.02\text{e-}22\ \ 1.07\text{e-}21\ \ 1.07\text{e-}21\ \ 1.0\ \ 9.50\text{e-}21\ \ 1.07\text{e-}21\ \ 1.07\text{e-}21\ \ 1.0]^T\\
h_7 &= [2.77\text{e-}25\ \ 3.44\text{e-}25\ \ 3.44\text{e-}25\ \ 1.0\ \ 1.39\text{e-}25\ \ 3.44\text{e-}25\ \ 3.44\text{e-}25\ \ 1.0]^T\\
h_8 &= [2.64\text{e-}30\ \ 1.1\text{e-}28\ \ 1.1\text{e-}28\ \ 1.0\ \ 2.64\text{e-}30\ \ 1.1\text{e-}28\ \ 1.1\text{e-}28\ \ 1.0]^T\\
o_8 &= 0.024
\end{align*}
\begin{itemize}
\item
Input: $w = (_1)_1(_2(_1)_1$
\end{itemize}
For this example we obtain
\begin{align*}
h_0 &= [2.4\text{e-}02\ \ 1.0\ \ 1.0\ \ 1.0\ \ 2.4\text{e-}02\ \ 1.0\ \ 1.0\ \ 1.0]^T\\
h_1 &= [1.30\text{e-}04\ \ 3.20\text{e-}04\ \ 3.2\text{e-}04\ \ 1.0\ \ 2.58\text{e-}04\ \ 3.20\text{e-}04\ \ 3.20\text{e-}04\ \ 1.0]^T\\
h_2 &= [2.46\text{e-}09\ \ 1.02\text{e-}07\ \ 1.02\text{e-}07\ \ 1.0\ \ 2.48\text{e-}09\ \ 1.02\text{e-}07\ \ 1.02\text{e-}07\ \ 1.0]^T\\
h_3 &= [2.64\text{e-}11\ \ 3.28\text{e-}11\ \ 3.28\text{e-}11\ \ 1.0\ \ 1.33\text{e-}11\ \ 3.28\text{e-}11\ \ 3.28\text{e-}11\ \ 1.0]^T\\
h_4 &= [5.88\text{e-}15\ \ 1.05\text{e-}14\ \ 1.05\text{e-}14\ \ 1.0\ \ 9.24\text{e-}15\ \ 1.05\text{e-}14\ \ 1.05\text{e-}14\ \ 1.0]^T\\
h_5 &= [2.70\text{e-}18\ \ 3.36\text{e-}18\ \ 3.36\text{e-}18\ \ 1.0\ \ 1.36\text{e-}18\ \ 3.36\text{e-}18\ \ 3.36\text{e-}18\ \ 1.0]^T\\
o_5 &= 0.805
\end{align*}
\begin{itemize}
\item
Input: $w = (_2)_1(_1)_2)_2$
\end{itemize}
For this example we obtain
\begin{align*}
h_0 &= [2.4\text{e-}02\ \ 1.0\ \ 1.0\ \ 1.0\ \ 2.4\text{e-}02\ \ 1.0\ \ 1.0\ \ 1.0]^T\\
h_1 &= [2.58\text{e-}04\ \ 3.20\text{e-}04\ \ 3.20\text{e-}04\ \ 1.0\ \ 1.30\text{e-}04\ \ 3.20\text{e-}04\ \ 3.20\text{e-}04\ \ 1.0]^T\\
h_2 &= [2.07\text{e-}07\ \ 1.02\text{e-}07\ \ 1.02\text{e-}07\ \ 1.0\ \ -2.02\text{e-}07\ \ 1.02\text{e-}07\ \ 1.02\text{e-}07\ \ 1.0]^T\\
h_3 &= [2.64\text{e-}11\ \ 3.28\text{e-}11\ \ 3.28\text{e-}11\ \ 1.0\ \ 1.33\text{e-}11\ \ 3.28\text{e-}11\ \ 3.28\text{e-}11\ \ 0.0]^T\\
h_4 &= [2.52\text{e-}16\ \ 1.05\text{e-}14\ \ 1.05\text{e-}14\ \ 1.0\ \ 2.52\text{e-}16\ \ 1.05\text{e-}14\ \ 1.05\text{e-}14\ \ 0.0]^T\\
h_5 &= [-1.30\text{e-}17\ \ 3.36\text{e-}18\ \ 3.36\text{e-}18\ \ 1.0\ \ -6.31\text{e-}18\ \ 3.36\text{e-}18\ \ 3.36\text{e-}18\ \ 0.0]^T\\
o_5 &= 4.88
\end{align*}

\end{document}